\documentclass{article}
\usepackage[margin=1in]{geometry}
\usepackage{graphicx} 
\usepackage{amsmath}
\usepackage{amssymb}
\usepackage{mathtools}
\usepackage{comment}
\usepackage{todonotes}
\usepackage[all, rotate]{xypic}
\usepackage{ulem}
\usepackage{hyperref}
\usepackage{authblk}

\usepackage{biblatex} 
\title{Document Understanding, Measurement, and Manipulation\\Using Category Theory}
\author[1]{Jared Claypoole}
\author[1]{Yunye Gong}
\author[2]{Noson S. Yanofsky}
\author[1]{Ajay Divakaran}
\affil[1]{SRI International}
\affil[2]{Brooklyn College}

\date{October 2025}

\begin{document}

\maketitle

\begin{abstract}
We apply category theory to extract multimodal document structure which leads us to develop information theoretic measures, content summarization and extension, and self-supervised improvement of large pretrained models. We first develop a mathematical representation of a document as a category of question-answer pairs. Second, we develop an orthogonalization procedure to divide the information contained in one or more documents into non-overlapping pieces. The structures extracted in the first and second steps lead us to develop methods to measure and enumerate the information contained in a document. We also build on those steps to develop new summarization techniques, as well as to develop a solution to a new problem viz. exegesis resulting in an extension of the original document. Our question-answer pair methodology enables a novel rate distortion analysis of summarization techniques.
We implement our techniques using large pretrained models, and
we propose a multimodal extension of our overall mathematical framework.
Finally, we develop a novel self-supervised method using RLVR to improve large pretrained models using consistency constraints such as composability and closure under certain operations that stem naturally from our category theoretic framework.
\end{abstract}



\section{Introduction}

Our motivation is to extract structure from documents as a means for manipulating them.  
Category theory\cite{yanofsky_monoidal} is the modern way to describe structure in mathematics and in the sciences. We use methods based on category theory to describe the structure and information content of a document. We can represent any document as a category of question-answer pairs the document answers.
We will explore the benefits of extracting such structure throughout this work.
With this structure in hand, we can go on to describe summaries and extensions of documents. 
We give several measures of information and entropy of a document. 

We are very general in our definition of a document.
We can mean text, images, audio, video, raw sensor data, or other modalities, or a mix of multiple modalities.
Our work applies to all of them and hence we are modality agnostic.
While much of our present work uses examples from the text modality, we later extend to describing high-level semantics of multimodal documents using the same structures.

One can view this work as using the following theories to work on documents. This forms our ever-expanding toolbox of manipulation and measurement tools.

\begin{equation*}
    \xymatrix{
        *+[F]{\txt{Rhetorical structure\\ theory}}
        \ar[dr]
        & *+[F]{\txt{Category theory and \\Homotopy theory}}
        \ar[d]
        & *+[F]{\txt{Information\\theory}}
        \ar[dl]
        \\
        & *+[F]\txt{Documents}
        \ar[dd]^*+[F.]{\txt{Large pretrained model\\instruction following}}
        \\
        \\
        & *+[F]\txt{Toolbox}
    }
\end{equation*}

Our approach extends Shannon’s foundational work on syntactic information: whereas Shannon focused on the statistical properties of symbol sequences, we apply similar principles to the semantic layer, by discretizing meaning itself.
We will ultimately use large pretrained models as a means of extracting latent semantic structures from documents.
Large pretrained models are the necessary tools that Shannon and others did not have; they are key to our making concrete progress.

Discretization here refers to the transformation of context-dependent, natural language semantic representations into structured, manipulable units. Once semantic information is discretized, it becomes amenable to measurement, comparison, and transformation. This enables us to quantify semantic relationships, track shifts in meaning, and even extend reasoning processes over abstract concepts.

In effect, large pretrained models act as semantic compressors: they reduce the complexity of human meaning into tractable forms, allowing us to probe, reshape, and analyze the informational content of a document in ways that were previously inaccessible.

The following is a graphical representation of our workflow for representing, manipulating, and measuring documents.
\begin{align*}
    \xymatrix{
        & & & & *+[F-,]\txt{Suppression/\\Summaries}
        & *+[F-,]\txt{Exegesis/\\Extensions}
        \\
        *+[F]\txt{Raw\\document}
        \ar[r]
        & *+[F]\txt{Abstractive\\DAG}
        \ar[r]
        & *+[F]\txt{DAG of\\core QAs}
        \ar[r]
        & *+[F]\txt{Category\\ of QAs}
        \ar[r]
        \ar[d]
        & *+[F]\txt{Orthogonalized\\QAs}
        \ar[r]
        \ar[d]
        \ar[dr]
        \ar@/^1.2cm/[dd]
        \ar[ddr]
        \ar[ddl]
        & *+[F]\txt{Lattice\\of QAs}
        \ar[ul]
        \ar[u]
        \\
        & & & *+[F-:<6pt>]\txt{Content\\entropy}
        & *+[F-:<6pt>]\txt{Information\\content}
        & *+[F-:<6pt>]\txt{Mutual\\information}
        \\
        & &
        & *+[F-:<6pt>]\txt{Diversity\&\\depth\\entropy}
        & *+[F-:<6pt>]\txt{Information\\density}
        & *+[F-:<6pt>]\txt{Information\\gain}
    }
\end{align*}

Let us describe this workflow in more detail.
\begin{itemize}
    \item We begin with a raw document.  This could be any modality or mixture thereof.
    \item We then extract the rhetorical structure of the document into what we call an abstractive DAG.
    \item Next we convert this rhetorical structure into question-answer pairs that retain all the information from the abstractive DAG nodes, organized in a DAG of core question-answer pairs.
    \item From that we represent the document as a category of question-answer pairs that contains the core QAs as well as additional QAs that have been decomposed from them.
    \item Next we define a metric and perform an orthogonalization procedure, identifying atomic QA pairs that are non-overlapping and that span the space of QAs derived from the document.
    \begin{itemize}
        \item From the category of QAs and the set of orthogonalized QAs, we define many measures of the entropy or information content of documents.
    \end{itemize}
    \item Finally, we construct a lattice of QAs from the set of orthogonalized QAs and the underlying rhetorical structure of the document, representing all possible subsets of the set of orthogonalized QAs in a hierarchical way.
    \begin{itemize}
        \item The hierarchical subsets represented by the lattice precisely correspond to summaries of the document.
        \item We can extend this lattice to perform exegesis, creating extensions of the document.
    \end{itemize}
\end{itemize}

Our major contributions are as follows:

\begin{itemize}
    \item A mathematical representation of information in a document in the form of a category of question-answer pairs
    \item An orthogonalization procedure to divide the information in one or more documents into non-overlapping pieces
    \item Methods to extract the rhetorical structure of a document
    \item Methods to measure and enumerate the information contained in a document in terms of the QAs answered by the document
    \item Methods to create and organize summaries and extensions of a document
    \item Methods to relate different documents
    \item Rate distortion analysis
    \item Extensions to multiple modalities
    \item Implementations of these ideas using large pretrained models
    \item Use of our structures to generate automated constraints that can be fed back to a model
\end{itemize}

We begin by discussing the core of our ideas, which is representing document structure using question-answer pairs, organizing and decomposing them in a category, then defining a metric and orthogonalization procedure.

We then move on to our methods of extracting and applying the rhetorical structure of a document, describing the workflow enumerated above in detail.

Finally we describe further directions and related work.

\section{QA structure representation}

We now describe the mathematical core of our ideas.

\subsection{Equivalence between assertions and question-answer pairs}
Let us begin with assertions.  We will consider two assertions to be equivalent using question answering: for any question-answer pair one assertion is able to answer, the equivalent assertion must be able to answer it consistently with the first.

Throughout this work we will rely heavily on question-answer pairs (QA pairs, or QAs). These QAs are in direct correspondence with core assertions, which are defined as restatements of each question-answer pair as an assertion.
We can also take any assertion and construct core QA pairs from it.  A core QA corresponding to an assertion is one whose restatement as an assertion is equivalent to the original assertion, in the question-answering sense defined above.
That is, the restated assertion can answer any question that the original statement could answer.

The mapping between questions and assertions is not unique, but the mapping between their equivalence classes is unique.  That is, any assertion uniquely maps to an equivalence class of core QAs, and vice versa.

We denote equivalence classes with square brackets.
If $QA_1$ is equivalent to $QA_2$, we say $[QA_1] = [QA_2]$.
We denote the core assertion to $QA_1$ as $\mathrm{core}(QA_1)$ and the class of questions whose core assertions are equivalent to assertion $A_1$ as $[\mathrm{core}(A_1)]$.  As stated above, $\mathrm{core}(A_1)$ is itself not uniquely defined, because there are in general many question-answer pairs corresponding to a given assertion, but $[\mathrm{core}(A_1)]$ is well defined.  Thus if $\mathrm{core}(QA_1) = A_1$,
$[A_1]$ is isomorphic to $[\mathrm{core}(QA_1)]$.

\subsection{A category of assertions and question-answer pairs}

\subsubsection{Definition}

First an aside on categories.
Categories are collections of structures and ways of changing
those structures. The theory of categories has emerged as a
theory of structures and processes. As such, they have been used
to describe many different phenomena in mathematics and
science. A category can be thought of as a souped up graph in which in addition to connectivity between two nodes, there is also a notion of reachability from any node to any other node. Such reachability is a result of a specialized connectivity that is achieved through the definition of categories as follows. A category consists of objects and morphisms defined between those objects, with the morphisms required to support an identity operation as well as composability. Thus for any morphism ``$\to$,'' if $A\to B$ and $B\to C$, then $A\to C$ and $A\to A$. Note that the composability is what ensures the reachability we alluded to earlier.
While connectivity is a local property, reachability captures the global structure of the graph, which enables categories to tightly capture global and local structure. Category theory represents the state of the art in mathematical representation of structure.

Now let us construct a partial order category whose objects are equivalence classes of question-answer pairs.
Because they are isomorphic in the sense described above,
we consider each equivalence class of QA pairs to also be equivalent to an equivalence class of assertions: $[QA] = [\mathrm{core}(QA)]$.

Morphisms in this category represent question answering ability:  there exists a morphism from $[QA_1]$ to $[QA_2]$ if and only if $[\mathrm{core}(QA_2)]$ is able to answer $[QA_1]$.
Composition works because question answering is transitive:  if $\mathrm{core}(QA_2)$ can answer $QA_1$ and $\mathrm{core}(QA_3)$ can answer $QA_2$, then $\mathrm{core}(QA_3)$ can answer $QA_1$.
Identity morphisms are automatic, because $\mathrm{core}(QA_1)$ is always able to answer $QA_1$.

\subsubsection{Some examples}

Two general assertions $A$ and $B$, which may be related:
\begin{equation*}
    \xymatrix{
        & & A \cup B
        \\
        & A \ar[ur] & & B \ar[ul]
        \\
        A - B \ar[ur] & & A \cap B \ar[ul] \ar[ur] & & B - A \ar[ul]
    }
\end{equation*}

In the special case of two unrelated assertions, the diagram simplifies considerably:
\begin{equation*}
    \xymatrix{
        & A \cup B
        \\
        A \ar[ur] & & B \ar[ul]
    }
\end{equation*}

\begin{figure}[h!]
    \centering
    \includegraphics[width=0.4\textwidth]{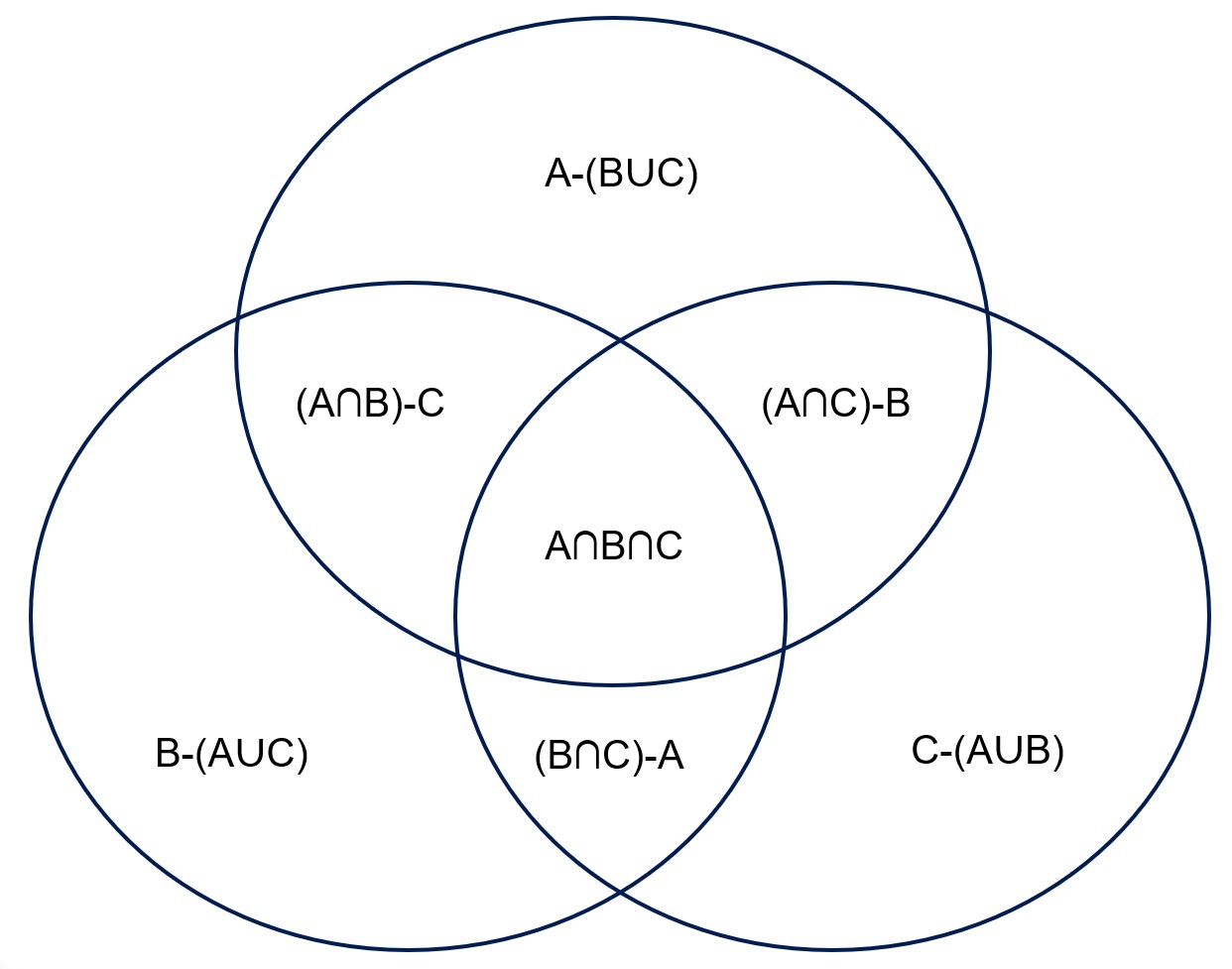}
    \hspace{0.05\textwidth}
    \includegraphics[width=0.4\textwidth]{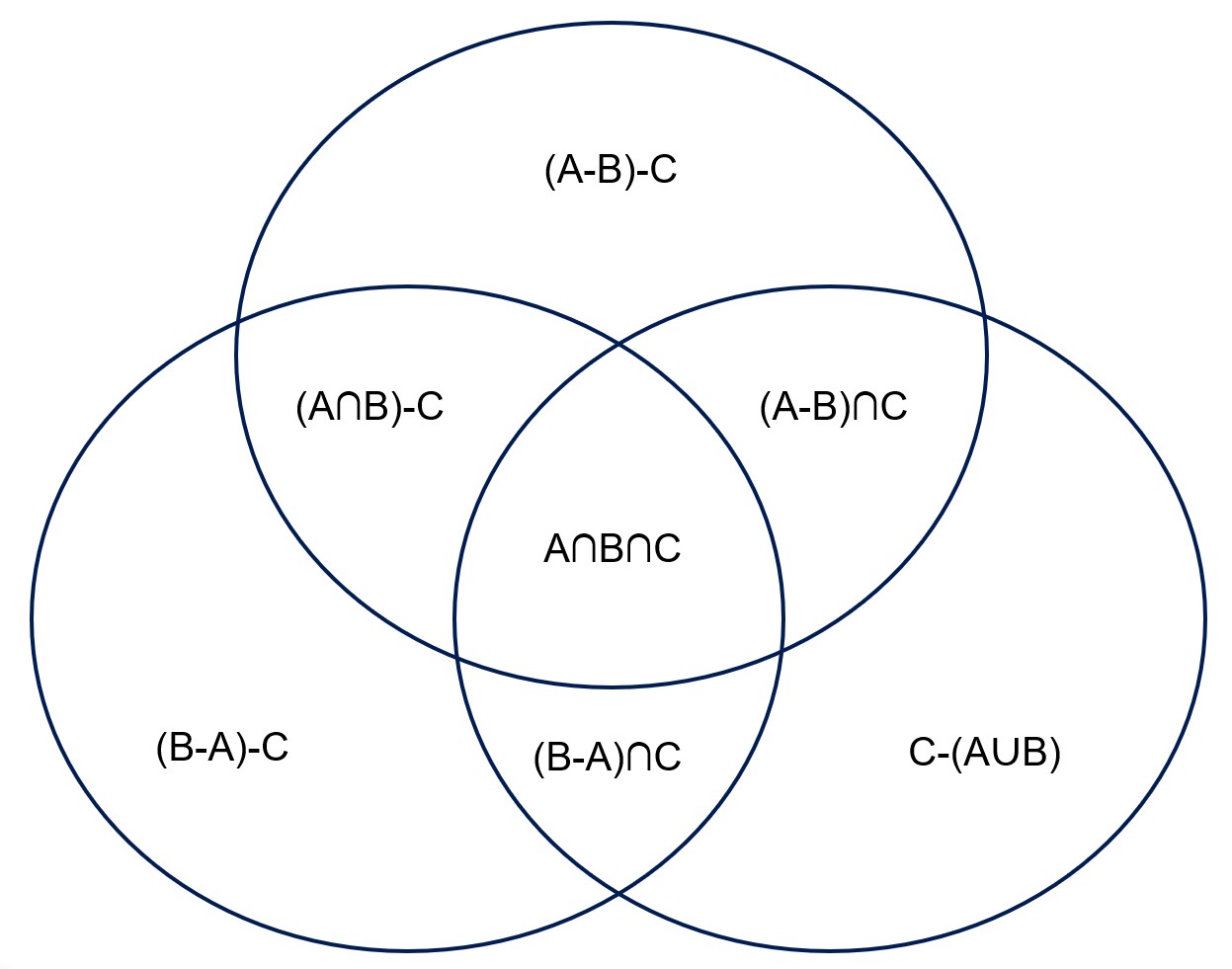}
    \caption{Two equivalent representations of non-overlapping components of three assertions $A$, $B$, and $C$.  On the left is a symmetric representation; on the right is a representation generated from $A$ and $B$ with $C$ then added.}
    \label{fig:three-sets-venn-diagram}
\end{figure}

Figure~\ref{fig:three-sets-venn-diagram} shows two equivalent representations of three related assertions $A$, $B$, and $C$.

\subsection{A metric and orthogonalization}

\subsubsection{Operations on QA pairs}

We seek to define unions, intersections, complements, and pair-wise decompositions of QA pairs.
At the end of the day, these concepts are not possible to operationalize or even perfectly define without some human-like judgement of what it means for two QA pairs to be overlapping or to contain the same information.
Later, we will operationalize these concepts using the judgements of large pretrained models.

We can take unions, intersections, and complements of QA pairs, and we can decompose QA pairs with respect to one another.  As we will see, the union operation can require the others in the case of inconsistent QA pairs, and the decomposition operation is intertwined with the intersection and complement operations.

When we decompose QA pairs with respect to one another, we seek to rewrite them in equivalent ways that preserve all the information present in the original QAs yet separate them into pieces that are either entirely overlapping or entirely non-overlapping.
The fundamental relationship between intersection, complement, and decomposition can be described as follows:
\begin{equation*}
    \mathrm{decomp}(QA_1, QA_2) = (QA_1 - QA_2, QA_1 \cap QA_2, QA_2 - QA_1)
\end{equation*}
That is, we decompose the two QAs into a piece that contains all the information present in $QA_1$ but not $QA_2$ (which we call $QA_1-QA_2$, the complement of $QA_1$ with respect to $QA_2$), a piece that contains all the information present in both $QA_1$ and $QA_2$ (which we call the intersection of the two), and the complement of $QA_2$ with respect to $QA_1$.  The combination of the three (formally, their union) is equivalent to the combination of the original pair of QAs.

The union of two QA pairs depends on the nature of the QAs.  If their core assertions are consistent with one another as logical propositions, then we combine them with a logical ``and.''
On the other hand, if the core assertions corresponding to the two QAs are logically inconsistent, we combine them with a logical ``or''.
In summary,
\begin{equation*}
    QA_1 \cup QA_2 = \left\{ \begin{matrix*}[l]
        QA_1 \land QA_2, \text{ if $\mathrm{core}(QA_1)$ and $\mathrm{core}(QA_2)$ are logically consistent propositions}
        \\
        QA_1 \lor QA_2, \text{ otherwise}
    \end{matrix*}\right.
\end{equation*}

Combining two consistent QA pairs with a logical ``and'' is simple: we simply combine the questions and the answers separately with ``and.''  So\\ $(\text{``what is the color of the sky?  (blue)''}) \cup (\text{``what is the color of the grass? (green)''})$ becomes ``what is the color of the sky and what is the color of the grass?  (blue and green, respectively).''

Combining two inconsistent QA pairs with a logical ``or'' is a bit more complex.\footnote{
When dealing with equivalence classes, the process is much simpler.  We can convert each QA pair to an assertion, combine the assertions with a logical ``or,'' and convert the compound assertion into a core QA pair.  But this requires one to deal in equivalence classes, because for any assertion $A$, $\mathrm{core}(A)$ is not well defined, as there are many equivalent questions that can be made that contain all the information in $A$.  However, $[\mathrm{core}(A)]$ is well defined.
}
First one needs to decompose the QA pairs into pieces that isolate the inconsistency as two different answers to the same question plus pieces that are consistent.  Then combine the consistent pieces as usual (with logical ``and''s) and combine the two different answers to the same question with a logical ``or,'' and finally combine the resulting pieces with logical ``and''s.
For example, $(\text{``what is the color of the water and the sky?  (blue)''}) \cup (\text{``are the sky and the clouds both gray (yes)''})$ decomposes into ``what is the color of the water, what is the color of the clouds, and what is the color of the sky? (blue, gray, and blue or gray, respectively).''\footnote{
One could object that the form of the second question, ``are the water and the clouds both gray,'' was changed during this operation.
This is unavoidable when dealing with inconsistent QAs in general.  If one desires, one could define these operations purely on equivalence classes of QAs, in which case they would be perfectly well defined.
We do not take this approach because we believe it is useful to perform the decomposition in a way that  preserves the structure of the QAs reasonably well, a distinction that is lost when dealing with equivalence classes.
}

In practice, we will use large pretrained models to conduct these operations.

\subsubsection{A metric between assertions and QA pairs}

We define the following metric between two assertions (or, equivalently, between two associated core question answer pairs):
\begin{equation*}
    d(A, B) = 1 - \frac{\text{Number of QAs both assertions can answer}}{\text{Number of QAs either assertion can answer}}
    = 1 - \frac{|\mathcal{QA}(A)\cap \mathcal{QA}(B)|}{|\mathcal{QA}(A)\cup \mathcal{QA}(B)|},
\end{equation*}
where $\mathcal{QA}(A)$ is the set of question-answer pairs that can be answered by assertion $A$.
For QAs, we have $d(QA_1, QA_2) = d(\mathrm{core}(QA_1), \mathrm{core}(QA_2))$

For example, consider $A=\text{``the sky is blue''}$ and $B=\text{``the sky is not blue''}$.
We generate the questions $\mathcal{QA}(A,B)=\{QA_1, QA_2\}$, with $QA_1=\text{``what color is the sky? (blue)''}$
and $QA_2=\text{``is the sky blue? (yes)''}$.
Assertion $A$ can answer both questions, while assertion $B$ can answer only the second; thus the distance between them is $d(A,B)=\frac{1}{2}$.

This is an example of the Jaccard metric, which satisfies the triangle inequality.
See Gilbert's terse yet simple proof~\cite{gilbert_1972_jaccard-triangle-inequality}.

Note that this metric is symmetric, and $d(A,A)=0$ for any assertion $A$.
We will consider equivalent any assertions that answer exactly the same questions, so $d(A,B)=0$ implies $A$ is equivalent to $B$.
Two orthogonal assertions will have the maximum distance of 1 from one another.

\subsubsection{A metric between documents}

We can extend the previous metric to the case of comparing two different documents.  Each document has an associated category of QAs, and computing the metric will involve merging the categories and merging QAs that are isomorphic to one another.

\begin{equation*}
    d(D_1, D_2) =
    1 - \frac{\text{Number of merged QAs in the merged category}}{\text{Total number of QAs in the merged category}}
    = 1 - \frac{|\mathcal{QA}(D_1)\cap\mathcal{QA}|(D_2)|}{|\mathcal{QA}(D_1)\cup\mathcal{QA}(D_2)|}
\end{equation*}

\subsubsection{Orthogonalization}\label{sec:ortho-procedure}

We orthogonalize a category of QA pairs into a category of ``atomic'' QA pairs as follows.
First we compare each pair of QAs and decide whether they have any overlap with one another.  (If they have overlap, they will be able to answer questions in common, and thus would not be considered orthogonal by our metric.)
If the QAs have overlap, we decompose them into component pieces that are either completely equivalent or completely orthogonal.

Here is the detailed procedure to orthogonalize.  Start with an empty pool of already processed QAs, and choose an initial QA to place in it.  
(Note that ``already processed QAs'' might still be decomposed later in the process.)
Then iteratively choose another QA to process next.

When processing a QA, atomize it with respect to every other QA in the already processed pool.  For each pair of QAs, if either QA decomposes, replace the original QAs with their decomposed versions.
Crucially, QAs in the already processed pool can be decomposed and replaced with their decomposed versions.

The idea is to compare every QA to every other QA, decomposing away any overlap.
We postulate that if $QA_1$ has no overlap with $QA_2$, then decomposed versions of $QA_1$ will have no overlap either with $QA_2$ or QAs decomposed from $QA_2$.
\footnote{This lack of overlap could be used as a consistency condition.  See Section~\ref{sec:constraints} for more details.}

\section{Extracting and applying rhetorical structure}

Now we discuss our methods of extracting and applying the rhetorical structure of a document, describing our workflow in detail.

\subsection{Extracting rhetorical structure and the abstractive DAG}

Rhetorical structure theory is about extracting rhetorical structure from documents.
See \cite{rst_review_2020} for a survey of techniques, 
as well as \cite{liu-etal_2021_dmrst} and \cite{maekawa-etal_2024_rst-parsing-using-LLMs} for examples of newer language-model based techniques.
We use large pretrained models to extract rhetorical structure of documents.
We are also abstractive rather than extractive.

We use an {\bf abstractive DAG}\footnote{The current implementation of the abstractive DAG is a tree.  However, we choose to leave open the possibility that nodes can have multiple parents, as a given chunk of the document can rhetorically support multiple parent chunks.} to represent the rhetorical structure of a document.
Each node is a one sentence summary of a contiguous chunk of the document.
The root node is a one sentence summary of the entire document.
The leaf nodes are one sentence summaries of single sentences of the document, or even of partial sentences if it makes sense to further decompose a compound sentence.
When constructing the abstractive DAG in practice, we use a large pretrained model first to determine how to break the document into several semantically meaningful chunks at each stage and second to create a one sentence summary of each chunk.
The different levels of the abstractive DAG each comprise a complete summary of the document, each at a different level of abstraction.
Because these summaries are abstractive rather than extractive,\footnote{An extractive summary takes pieces of text verbatim from the original document. An abstractive summary paraphrases the meaning of the original document.} they are able to make inferences implied by but not directly stated in the original document.


\subsection{Creating a category of QAs from a abstractive DAG}

Starting from the abstractive DAG, we convert each node's assertion into one or more core QA pairs that capture all the information present in the assertion, resulting in a {\bf DAG of core QAs} with the same structure as the abstractive DAG.
In cases where the assertion is a compound statement and it is not natural to construct a single QA pair, we first decompose the assertion into sub-statements whose union is equivalent to the original assertion, and we then construct a core QA pair from each sub-statement.

Next we decompose the core QAs into sub-assertions/ sub-QAs, giving us the skeleton of a {\bf category of QA pairs}.
We then fill in the skeleton by performing a decomposition procedure (as described in Section~\ref{sec:ortho-procedure}) where each pair of QAs gets decomposed into pieces that are either entirely overlapping or entirely non-overlapping.  Throughout this decomposition procedure, we add question-answering morphisms from decomposed QAs to their parents, and we identify isomorphisms between entirely overlapping QAs.

Moreover, we retain the structure of the abstractive DAG in this category of QA pairs.
This can be understood as a function\footnote{
Actually, this function $f:\mathrm{obj}_\mathbb{C} \to \mathcal{P}(\mathrm{nodes}(\text{abs-dag}))$ (where $\mathbb{C}$ is the category of QA pairs and $\mathcal{P}(\mathrm{nodes}(\text{abs-dag}))$ is the power set of nodes of the abstractive DAG)
is better described as a total-valued, surjective relation $\tilde{f}: \mathrm{obj}_\mathbb{C} \to \mathrm{nodes}(\text{abs-dag})$.
That is, each QA pair in $\mathbb{C}$ is related to one or more nodes of the abstractive DAG.
Then the inverse relation is perfectly well defined, and in fact will also be a total-valued, surjective relation.
See Section~\ref{sec:relations} or \cite{yanofsky_monoidal} for more details about relations.
} that maps each QA in the category to the set of abstractive DAG nodes from which it was constructed or decomposed. There is also a notion of an inverse function that maps each node of the abstractive DAG to the QAs that were constructed or decomposed from it.

\begin{figure}
    \centering
    \includegraphics[width=\textwidth]{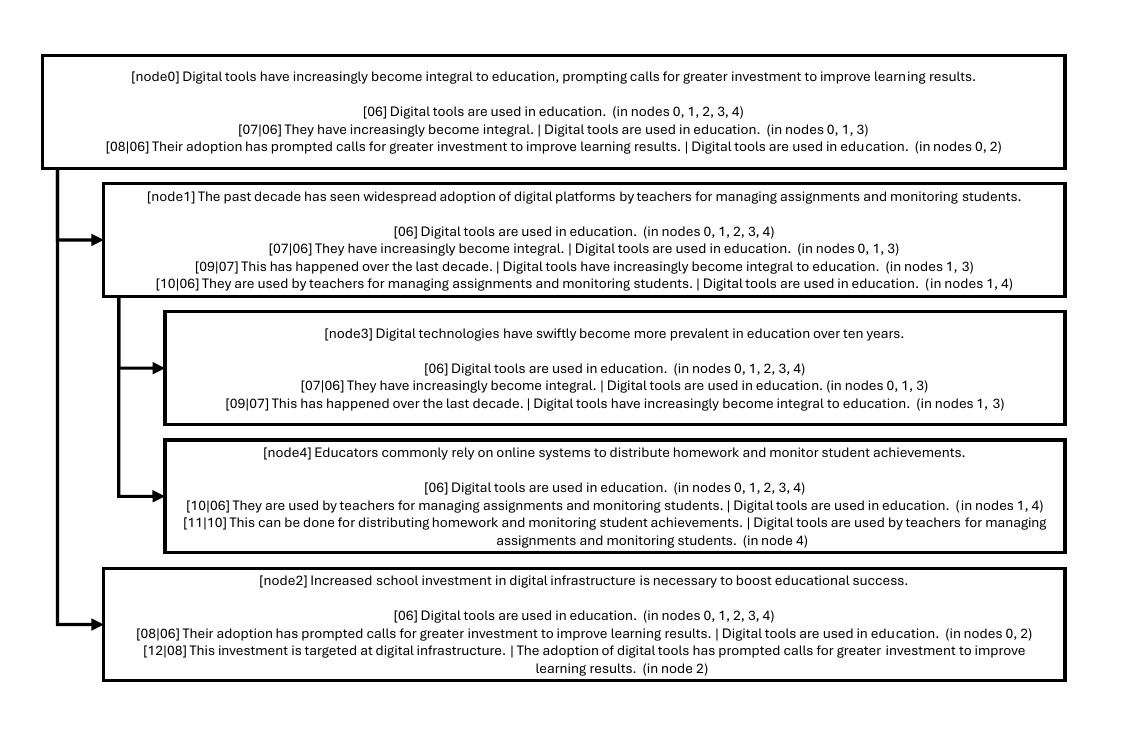}
    \caption{
        An example of a DAG of orthogonalized QAs.
        Here we represent the QAs using their core assertions.
        Each node represents a node of the abstractive DAG, with its original assertion on the top line.
        The following lines are the decomposed, orthogonal QAs.
        This DAG was created from the following sentences:\\\\
        ``The integration of digital tools in education has grown rapidly over the past decade. Many teachers now use online platforms to assign homework and track student progress. Schools should invest more in digital learning infrastructure to enhance educational outcomes.''
    }
    \label{fig:ortho-dag-edu}
\end{figure}

Finally, we construct a set of {\bf orthogonalized QA pairs} from the category of QA pairs, using the procedure described in Section~\ref{sec:ortho-procedure}.
This construction retains the rhetorical structure using the same function as the category of QA pairs, restricted to the set of orthogonalized QA pairs.
This mapping allows us to think of the set as a DAG of orthogonal QA pairs.
Note that the notion of the restricted inverse is well defined, because each node of the DAG will map to at least one orthgonal QA pair.
Figure~\ref{fig:ortho-dag-edu} is an example of a DAG of orthogonalized QAs, represented using their core assertions.

In practice we represent complements of assertions using a complement operator (``$|$'').
We interpret $A|B$ as ``If $B$ were to be the case, then $A$ would be the case.''
We can think of it as saying $A$ is conditioned on $B$.
The point of this is to remove the information present in $B$ from $A$.
We need this operator because it is often impossible to completely orthogonalize natural language statements without it,
as natural language statements require a certain amount of context in order to make sense.
Consider the statements ``I ate a roll while walking down the street'' and ``I ate a roll to fill myself up.''
The intersection of the two is clearly ``I ate a roll,'' but what remains of each of the two statements after removing their intersection?
We solve this problem by writing e.g., ``I did so to fill myself up $|$ I ate a roll.''
\footnote{
    We choose generally to write the first part of $A|B$ using pronouns which refer to the condition ($B$).
    We find this reads better and is less clunky than something like ``I ate a roll to fill myself up $|$ I ate a roll.''
}

\subsection{Lattices, summaries, and extensions} \label{sec:summ}

Exegesis is defined as ``the critical explanation or interpretation of a text, especially of scripture.''\footnote{Source: Oxford languages}
We consider exegesis to be a dual problem of summarization: in summarization, we suppress information from the original document; when performing exegesis, we add information to the original, obtaining an extension of the original document.  (We treat suppression and exegesis as the verbs and summaries and extensions as the resulting nouns.)
The following is a visual example of suppression (summary) and exegesis (extension):


\begin{align*}
    \xymatrix@=10pt{
        & A
        \\
        B \ar[ur] & & C \ar[ul]
        \\
        & D \ar[ur] & & E \ar[ul]
        \\
        & \txt{(a)}
    }
    & &
    \xymatrix@=10pt{
        & A
        \\
        B \ar[ur] & & *+[F--]{{C}} \ar@{-->}[ul]
        \\
        & *+[F--]{D} \ar@{-->}[ur] & & *+[F--]{E} \ar@{-->}[ul]
        \\
        & \txt{(b)}
    }
    & &
    \xymatrix@=10pt{
        & & A
        \\
        & *+[F]{U\cup B \cup V} \ar[ur] & & C \ar[ul]
        \\
        *+[F]{U} \ar[ur] & B \ar[u] & *+[F]{V} \ar[ul] & D \ar[u] & E \ar[ul]
        \\
        & & \txt{(c)}
    }
\end{align*}
In the diagram above, (a) is a representation of an example document with five assertions; (b) is a representation of suppression of assertion $C$ and its children ($D$ and $E$), where the dashed elements are removed, resulting in a summary; and
(c) is a representation of exegesis: adding assertions $U$ and $V$, modifying the structure of the document in the process, and resulting in an extension of the original document.

The orthogonalized QAs, together with their map back to the abstractive DAG structure, generate a hierarchical lattice that can be used to represent summarization.
The lattice is pictured below (right) for a simple document (left) consisting of only three assertions, with assertion $A$ higher in the hierarchy than assertions $B$ and $C$.

\newcommand{\settup}[2]{{ (\{#1\}, \{#2\}) }}

\begin{align*}
    \xymatrix{
        \\
        & A
        \\
        B \ar[ur] & & C \ar[ul]
    }
    & &
    \xymatrix{
        &
        *+[F]{\settup{A}{B, C}}
        &
        \\
        *+[F]{\settup{A}{B}}
        \ar[ur]
        &
        *+[F]{\settup{A}{C}}
        \ar[u]
        &
        *+[F--]{\settup{}{B,C}}
        \ar@{-->}[ul]
        \\
        *+[F]{\settup{A}{}}
        \ar[u]
        \ar[ur]
        &
        *+[F--]{\settup{}{B}}
        \ar@{-->}[ul]
        \ar@{-->}[ur]
        &
        *+[F--]{\settup{}{C}}
        \ar@{-->}[ul]
        \ar@{-->}[u]
        \\
        &
        *+[F]{\settup{}{}}
        \ar[ul]
        \ar@{-->}[u]
        \ar@{-->}[ur]
    }
\end{align*}

The lattice above (right) expresses all the possible ways to summarize the document $D$ whose rhetorical structure is shown on the left.  At the bottom is the empty summary, and at the top is the entire original document.  In the middle two rows are versions with some but not all of the assertions suppressed.
Each node of the lattice is of the form $(S_1, S_2)$, where $S_1$ and $S_2$ are sets of assertions from the top and bottom levels of the hierarchy respectively.
We call a summary hierarchical if it respects the rhetorical structure of the document, i.e., it doesn't suppress assertions higher in the rhetorical structure before suppressing assertions lower in the rhetorical structure.  In this case, a hierarchical summary must preserve assertion $A$ before preserving assertions $B$ and $C$.  Nodes on the lattice corresponding to hierarchical summaries have solid outlines, and those that do not correspond to hierarchical summaries have dashed outlines.

We can decompose all summaries into a composition of two different types: quotient document summaries and subdocument summaries.
We can similarly decompose all extensions into a composition of two different types:
superdocument extensions and elaboration extensions.

When we take into account the categorical information of a document, we can discuss two different types of summaries and two different types of extensions. Keep an eye on the next page where we show examples of such summaries and extensions with the first few sentences of Abraham Lincoln's famous Gettysburg Address. 
The original document is in the center of the page. The two summaries are on the horizontal axis and the two extensions are
on the vertical axis. 

Summaries are documents which have only some of the information from the document. The rest of the document is crossed out. There is always an inclusion from a summary to the original document. 
\begin{itemize}
    \item A subdocument summary comes from crossing out entire chains of information. When the head of a chain is crossed out, all of its children must also be crossed out. They are giving more information about something that no longer exists. 
    \item A quotient document summary, on the other hand, crosses out parts of information that are on a chain. What remains are the main points. It has some of the main points of information. In the example below, exactly how many years ago was our nation born (``Four score and seven'') is not a main point. In contrast, a main point is ``years ago our fathers brought forth a new nation.'' Again, we ignore details about why the nation was brought forth (``conceived in liberty, and dedicated to ...''). The reason why such a summary is called a quotient document summary is that there is a quotient map from the original document to such summaries. The QAs in the original document that are crossed out in the summary are related to the chains that are still in the summary. This is in stark contrast to subdocument summaries which do not have functions from the original document. 
\end{itemize}

A typical summary will have parts that are of the form of a subdocument summary and parts that have the form of a quotient document summary.  

Extensions are documents that have the original document, and more. The extra parts are underlined.  
\begin{itemize}
    \item An elaboration extension adds details to parts that are already in the document. So in the address, we know why they are there, and we add in that not only ``our fathers'' but also ``our mothers'' brought forth this nation. We also add about when they came to field ``on this very pretty day.'' 
    
    \item A superdocument extension are formed by adding totally new heads of chains. For example, nowhere in the original document does it say where the field is.  So when we add in ``at Gettysburg, Pennsylvania'' we are adding in new information that is no where else. Similarly, ``After the speeches, a light lunch will be served'' is a totally new idea and not an elaboration of another idea.    
\end{itemize}

A typical extension will have parts which are like an  elaboration extension and parts that are like a superdocument extension. However, we have the following theorem: every extension can be written as a superdocument extension followed by a elaboration extension. The way to see this is that every document can first be extended by adding in all the new main points (heads of chains) and then further extensions by elaborating on all the chains (including the new ones.)


\begin{footnotesize}
\begin{equation*}
\xymatrix@C=0.5cm@R=0.5cm{
& {\parbox{7.5cm}{{\center \bf\large  Superdocument Extension\\added information}\\Four score and seven years ago our fathers brought forth on this continent, a new nation, conceived in Liberty, and dedicated to the proposition that all men are created equal. \underline{This new nation is better than most other nations} \underline{that were not dedicated to these ideas.}
Now we are engaged in a great civil war,  testing whether that nation, or any nation so conceived and so dedicated, can long endure. We are met on a great battle-field of that war \underline{at Gettysburg Pennsylvania}. We have come to dedicate a \underline{northern eastern corner} portion of that field, as a final resting place for those who here gave their lives that that nation might live. It is altogether fitting and proper that we should do this. \underline{After a few speeches, a light lunch will be served.}}} \\
{\parbox{4cm}{{\center \bf\large  Subdocument Summary\\some information}\\\sout{Four score and seven years ago} our fathers brought forth \sout{on this continent,} a new nation, conceived in Liberty, \sout{and dedicated to the proposition that all men are created equal.}
Now we are engaged \sout{in a great civil war,} testing whether that nation, \sout{or any nation so conceived and so dedicated}, can long endure. We are met on a great battle-field of that war. \sout{We have come to dedicate a portion of that field, }as a final resting place for those who here gave their lives that that nation might live. It is altogether fitting and proper that we should do this}} \ar@{^{(}->}[r] \ar@{^{(}->}[ru] \ar@{^{(}->}[rd]& 
{\parbox{5.5cm}{{\center \bf\large  Document}\\Four score and seven years ago our fathers brought forth on this continent, a new nation, conceived in Liberty, and dedicated to the proposition that all men are created equal.
Now we are engaged in a great civil war, testing whether that nation, or any nation so conceived and so dedicated, can long endure. We are met on a great battle-field of that war. We have come to dedicate a portion of that field, as a final resting place for those who here gave their lives that that nation might live. It is altogether fitting and proper that we should do this}} \ar@{->>}[r] \ar@{^{(}->}[u]\ar@<-3ex>@{^{(}->}[d] &
{\parbox{4cm}{{\center \bf\large  Quotient document Summary\\synopsis information}\\\sout{Four score and seven} years ago our fathers brought forth\sout{ on this continent,} a new nation, \sout{conceived in Liberty, and dedicated to the proposition that all men are created equal.}
\sout{Now} we are engaged in a\sout{ great civil} war, \sout{testing whether that nation, or any nation so conceived and so dedicated, can long endure.} We are met on a\sout{ great battle-}field of that war. \sout{We have come to dedicate a portion of that field, as a final resting place for those who here gave their lives that that nation might live.} It is \sout{altogether fitting and} proper that we should do this.}}\ar@{^{(}->}[lu] \ar@<-3ex>@{^{(}->}[l]  \ar@<-3ex>@{^{(}->}[ld] \\
& {\parbox{7.5cm}{{\center \bf\large  Elaboration Extension\\added details}\\Four score and seven years \underline{and three months} ago our fathers \underline{and mothers} brought forth on this continent \underline{between the Atlantic and the Pacific oceans}, a new nation, conceived in Liberty, \underline{prosperity} and dedicated to the proposition that all men are created equal.
Now we are engaged in a great \underline{and terrible} civil war, testing whether that nation, or any nation so conceived and so dedicated, can long endure. We are met on a great battle-field of that war. We have come to dedicate a portion of that field \underline{on this very pretty day}, as a final resting place for those who here gave their lives that that nation might live. It is altogether fitting and proper that we should do this.}} \ar@{->>}[u] \ar@{->>}[ur]
}
\end{equation*}
\end{footnotesize}

\subsection{Measures of document information}


Our framework allows us to measure document information in many, many ways.
Different measures will be useful for different purposes.
We introduce the following examples.

{\bf Information content} is defined to be the number of atomic orthogonal QA pairs associated with the document.
For a document $D$,
$IC(D) = |\mathcal{QA}(D)|$.

{\bf Information density} is information content normalized by some notion of the length of the document, such as the number of words, pages, etc.  $ID(D) = IC(D) / |D|$.

The {\bf mutual information} between two documents is the number of atomic orthogonal QAs in common between the documents (in the merged category).
This is related to but different from the similarity ($1-\text{dissimilarity}$) measured with the metric; here we do not normalize by the total number of QAs in the merged category, and here we orthogonalize before measuring.
$IC(D_1, D_2) = |\mathcal{QA}(D_1)\cap\mathcal{QA}|(D_2)|$.

The {\bf information gain} when supplying an additional document is given by
\begin{equation*}
    IG(D_2; D_1) = IC(D_2) - IC(D_1, D_2)
\end{equation*}
where $IC(D)$ is the information content of document $D$, and $IC(D_1, D_2)$ is the mutual information between documents $D_1$ and $D_2$.

For a given document $D$, the {\bf content entropy} of $D$, $CE(D)$, is the log of the number of ``chains'' in the category of QAs associated with $D$. A chain (or snake) is a 
sequence of QAs that point to one head. The main idea is that the head of the chain has all the information of the QAs in chain. The rest of the information is, in a sense, redundant. Another way to think of $CE(D)$ is as the logarithm of the number heads of chains. These are the major QAs that contain the information pointing to it. The measure is the logarithm of the number of QAs that are not the source of any non-trivial morphism, i.e., it is the head. 

The reason for the logarithm in the definition of $CE$ is that measure respects additivity. This means that $CE(D+D')=CE(D)+CE(D')$.  

In order to get a feel for the content entropy, consider the following three possible subcategories of QAs.   

\[
\xymatrix{
   & A & \\
   & B \ar[u] & \\
 C \ar[ur] &   & D \ar[ul]
}
\quad
\xymatrix{
  A & & B \\
  & C \ar[ul] \ar[ur] & \\
  & D \ar[u] &
}
\qquad
\xymatrix{
& A &\\
  & B\ar[u] & \\
  C \ar[ur] & & D \ar[ul] \\
  & E \ar[ul] \ar[ur] &\\
  &F\ar[u]
}
\]

The chain on the left has two tails. But this is considered only one chain. All the information is in the QA $A$. The main point is that the QA $B$ answers all the questions in $C$ and $D$. In contrast, the diagram in the center is considered to be two chains. Although $C$ has part of the information in $A$ and in $B$, there are two main pieces of information. The diagram on the right has one piece of information. The diamond in the center 
commutes because the category of QAs is a ``thin'' category, i.e., there is at most one morphism between any two objects. 

A few words must be said about calling this measure ``entropy.'' In Shannon's theory, a system with high entropy has low redundancy, since each symbol carries nearly maximal new information, whereas low entropy implies higher redundancy because patterns or repetitions reduce uncertainty. Here also, high entropy means that that there are a lot of heads of chains. This means that there are relatively few redundant QAs. Such a situation happens when the information is separated and does not interact. On the other hand, low entropy is when there are a lot of tightly interwoven QAs. Low entropy means a lot of redundancy. In a sense, the QAs that we are dealing with are analogous to the particles that Boltzmann was dealing with. If the particles are all jumbled together in one room, there is a low entropy. In contrast, if the particles are spread throughout the room and not near each other, then there is high entropy.

{\bf Content entropy density} for a document $D$ is the content entropy divided by the length of the document (e.g., the number of words or pages in the document). That is, $CEd(D)=CE(D)/|D|$.

Our combined {\bf diversity and depth entropy} measure involves performing a coding procedure on the DAG of atomic orthogonal QAs and counting the bits required to transmit the unique QAs.  We also have two slightly varied methods that can be treated as upper and lower bounds for this measure: the lower bound ($E_0$) involves coding all QAs independently, treating redundant QAs as distinct, and transmitting all QAs.  The upper bound ($E_1$) involves coding only unique QAs and transmitting each unique QA only once.
Our primary method ($E$) is a hybrid of the two: we code all QAs independently, pretending redundancies do not exist, then we transmit each unique QA only once.\footnote{When we speak of ``transmitting'' here, we simply refer to the number of bits that would be required to transmit these signals; no actual transmission is required.}

\subsection{Constraints and feedback} \label{sec:constraints}

Properties such as compositionality in our category of QA pairs give us constraints that we can apply to a model.
If the model fails to meet the constraints, we can perform fine tuning to obtain a model whose output better aligns with the mathematical properties we demand.
The DRESS system~\cite{chen_2024_dress} used reinforcement learning with AI feedback to improve model performance using a supervised fine tuning approach.
Another avenue for fine tuning is to take a DPO-like approach (as originally presented in~\cite{rafailov_2023_dpo}).

RLVR (Reinforcement Learning with Verifiable Rewards) \cite{Lambert_2025_rlvr}
has emerged as an effective way to provide automated feedback to large pretrained models post-training. It consists of using tasks with precisely computable results to provide corrective feedback to large models. It can be thought of as a self-supervised approach to improving large models since it does not use any human intervention. Our category theoretic framework naturally yields such tasks through constraints such as composability and closure under operations such as summarization. 
The composability constraint can be thought of as a mathematical generalization of consistency constraints. Since our summarization is derived from a lattice, it is closed under intersections and unions. For example, the union of two summaries is also a summary. Such constraints provide mathematical functions that can be automatically computed and thus provide a basis for self-supervision. Note that it is straightforward to generate a large number of such constraints automatically through the graph structure inherent in the category theoretic formulation. In forthcoming work, we will explore this avenue more thoroughly to get experimental results on improvements in large model performance accuracy.

\subsection{The category of documents} \label{cat-of-docs} \label{sec:relations}

The objects in the category of documents are the individual categories associated with instances of documents.
The morphisms are functors between document categories.
Exegesis and suppression are special examples of such functors.

Morphisms between documents need to be described as relations, where a relation between two sets $S$ and $T$ is a subset of $S\times T$, or a set of elements each of the form $(s, t)$, where $s\in S$ and $t\in T$.  A relation is a more general version of a function, where each element of the domain can be related to zero, one, or multiple elements of the codomain.
Relations can be represented as binary matrices, and composition of relations corresponds to binary matrix multiplication.
So if documents $D_1$ and $D_2$ contain $m$ and $n$ QAs respectively, then the relation $f:D_1 \to D_2$ will be described by an $n$ by $m$ matrix of binary values, specifying whether the $i$th QA of $D_2$ is related to the $j$th QA of $D_1$.
The main point is that each QA of one document may be related to zero, one, or more QAs in the other document. 

The morphism resulting from relational composition is possibly a sub-optimal morphism -- that is, if we have relations $f: D_1\to D_2$ and $g: D_2\to D_3$, $h=g\circ f$ could be sub-optimal in the sense that there may exist $h': D_1 \to D_3$ such that more assertions from $D_1$, $D_3$, or both are present in $h'$ than in $h$.  In other words, $h$ has too many assertions map to or from nothing.

The limit of sub-optimal morphisms are naive morphisms.  These 
exist between any two documents: one can relate any two documents by mapping all source assertions to nothing and having all target assertions come from nothing.
The interesting morphisms in this category are ones that map at least some assertions in document $D_1$ to assertions in document $D_2$.



\subsection{Rate distortion analysis}

\begin{figure}
    \centering
    \includegraphics[width=0.8\textwidth]{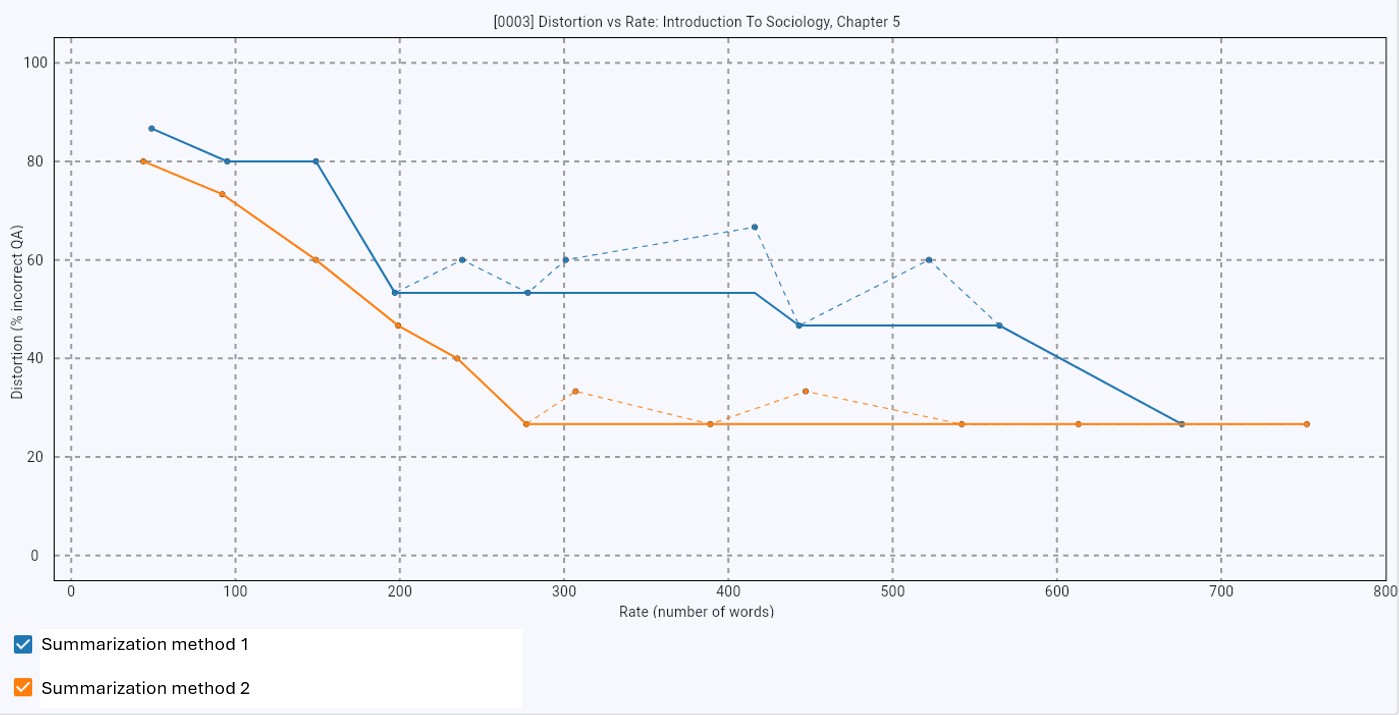}
    \caption{
        An example of rate distortion curves comparing two summarization methods. The rate is the number of words in the summary, and the distortion is the percentage of questions answered incorrectly.  The dashed lines connect the raw datapoints and the solid lines form the operational rate distortion curve, which uses the lowest distortion method at the given rate or lower.
    }
    \label{fig:rd-curve}
\end{figure}

Rate Distortion theory was originally applied to the problem of lossy compression (see Berger~\cite{berger_rd_book}) in which coding a source at a rate $R$ less than its entropy results in a distortion $D$. The rate distortion function $R(D)$ asserts that there exists a scheme that achieves the distortion $D$ using a rate $R(D)$, and that there is no scheme that can achieve distortion $D$ using less than $R(D)$ bits. In other words, $R(D)$ is the minimum rate required to achieve distortion $D$. In practice, compression schemes fall short of the ideal rate distortion function, but the rate distortion function provides a useful idea of how difficult or easy to compress a given source is. We can plot an ``operational'' rate distortion curve using a practical scheme to understand the tradeoff between rate and distortion.
We extend rate distortion theory to content summarization as follows. Let us assume that for a document $\mathcal{D}$, we have a question-answer set $\mathcal{QA}(\mathcal{D})$ that can be answered using the document $\mathcal{D}$. Then we summarize the document $\mathcal{D}$ to get the summary $\mathcal{S}$ as described earlier. Note that the suppression required to generate summaries (see Section~\ref{sec:summ}), is equivalent to lossy compression. The summary $\mathcal{S}$ can be used to answer a subset of the set of questions $\mathcal{QA}(\mathcal{D})$ and thus the percentage (or fraction) of questions it cannot help answer can be the distortion metric $D$. The rate $R$ can be the length of the summary in pages, words, lines, etc., as long as the definition is consistent. 
We can thus plot an operational rate distortion curve for a given summarization scheme by varying the length of the summary and calculating the percentage of questions not answered for each length. The operational rate distortion curve of a summarization scheme captures the trade-off between rate and distortion across the entire range of rates. We can thus compare summarization schemes by comparing their rate distortion curves. Please see Figure~\ref{fig:rd-curve} for an example.

\section{Further directions}
\subsection{Multimodal categories}

We extend the abstractive DAG to include additional modalities.
Its nodes continue to contain high level semantic assertions.  However, its nodes will now also point to low level evidence from additional modalities, such as text, images, video, audio, or other sensor data.
See the center of Figure~\ref{fig:cmder-intent-alignment} for a visual example of a multimodal category.


\subsection{Probabilistic categories}

We can add probabilistic uncertainty to assertions in the DAG structure as follows,
assigning a probability to each scenario.

\begin{equation*}
    \xymatrix@=15pt{
        & & & & {\bullet}\ar[dlll]_{p=0.3} \ar[d]^{p=0.2} \ar[drrr]^{p=0.5}
        \\
        & *+[F]\txt{Scenario 1} \ar[dl] \ar[d] \ar[dr] &
        & & *+[F]\txt{Scenario 2} \ar[dl] \ar[d] \ar[dr] &
        & & *+[F]\txt{Scenario 3} \ar[dl] \ar[d] \ar[dr] &
        \\
        QA_1 & QA_2 & QA_3
        & QA_1' & QA_2' & QA_3'
        & QA_1'' & QA_2'' & QA_3''
    }
\end{equation*}


\subsection{Coherent extensions}
There are many possible extensions one can give to a document. We can add lots of information to the information already available. However, we would be interested in making sure that the extension is somewhat coherent. This means that what was added into one part of the document does not conflict with what is added to another part of the document.

\begin{figure}[t]
    \centering
    \includegraphics[width=0.5\textwidth]{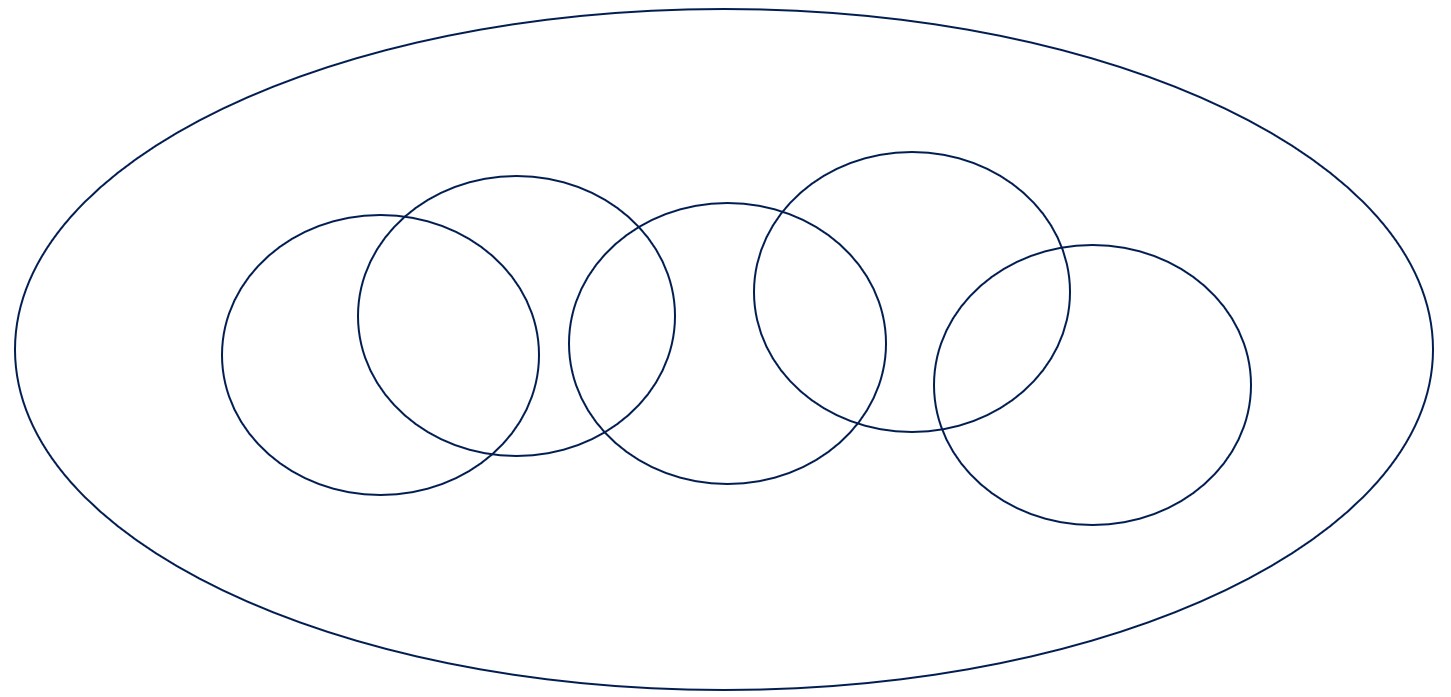}
    \caption{A topological space with open sets. A pre-sheaf adds information to each open set. If the information of the pre-sheaf matches on the overlap of open sets, then the pre-sheaf is actually a sheaf.}
    \label{fig:sheaf-circles}
\end{figure}

It is useful to think of extensions from the point of view of sheaves and pre-sheaves in algebraic topology. This abstract field of study is known for looking at local information of a 
topological space and determining global information about that space. This is done by assigning information to local open sets of the topological space (see Figure~\ref{fig:sheaf-circles}). Such an assignment is referred to as a pre-sheaf. If the information matches up on the overlap of any two open sets, then the assignment is called a sheaf. Using sheaves, one can capture certain types of global information about the space.

Recently sheaves have been used in much broader contexts. Rather than only looking at topological spaces, one can apply sheaves to more general combinatorial and mathematical problems~\cite{Rosiak2022Sheaf}. They are excellent for finding global solutions when there are local constraints.  

One can think of an extension of a document as adding information to the given document. Think of the different parts of the document as open sets. Then we look at different extensions to the different parts. We then ask how these different extensions fit together. The LLMs would have to look at the different parts and see if any QA from one part of the document contradicts another. More importantly, the LLMs will look if they agree with one extra QA extending the work. There are may possible results of this and there are many directions to go. 

Once we have coherent extensions understood, one can -- dare we say -- ``extend'' this work in many different ways. Here are some possibilities: 
\begin{itemize}
    \item Extending a coherent document with more information makes sense. However, it pays to think about what happens if we start with an incoherent document? That is start with a document $D$ that has parts that are inconsistent or have parts that imply inconsistencies. Does this mean we can add any information to the document and it will remain just as (in)consistent? This is similar to the logical idea of {\it ex falso quodlibet}, Latin for “from falsehood, whatever [you like].” This is also called ``Principle of Explosion.'' Or, perhaps there is some type of Paraconsistent Logic at play here which protects us from getting in trouble. 
    \item How is one to deal with many documents written by many different agents? Extensions of each of the documents are more prone to inconsistencies. Sheaf theory might also be very helpful in finding inconsistencies between documents rather than just looking at extensions of documents. We might be able to look at all the documents as one large topological space with each document being an open set of that topological space. From there, we can measure how far certain pre-sheaves are from being sheaves. This will correspond to a measure of how far many different but related documents are from being a unified whole. 
    \item Algebraic topology uses sheaves to encode local data and to study how it assembles into global invariants of a topological space. Although sheaf-theoretic methods can reflect aspects of dimension, such as through vanishing cohomology or support conditions, they do not directly determine the topological dimension of a space. We can ask to what extent local information about the document tells us about the ``dimension of a document.'' Along similar lines, we might be interested in ``hole'' in a document. What information is missing or is being avoided? What dimension are the ``holes'' in the document?  
\end{itemize}

These and other fascinating ways of going further have many application and connections to other areas. Much work remains.

\section{Discussion and outlook}

\subsection{Departure from the state of the art}

We rely on LLMs to access a much larger implicit ontology to discretize assertions, thus enabling a category theoretic representation of documents. The category theoretic representation combined with our distance metric enables projection of documents into a metric space. That in turn enables semantic retrieval as well as a systematic semantic merger of documents by merging their corresponding categories to create a semantically coherent super or mega-document. Our preliminary results on summarization indicate that our framework has empirical grounding. Our proposed document structure extraction thus enables a fine-grained understanding and manipulation of documents in semantic space. All in all our new framework leverages large pretrained models to make documents amenable to semantic and information theoretic analysis that was not possible until now.

\subsection{Potential applications}

Our framework offers a new paradigm for semantic document retrieval and manipulation that would enable the following potential applications.

\begin{figure}
    \centering
    \includegraphics[width=0.9\textwidth]{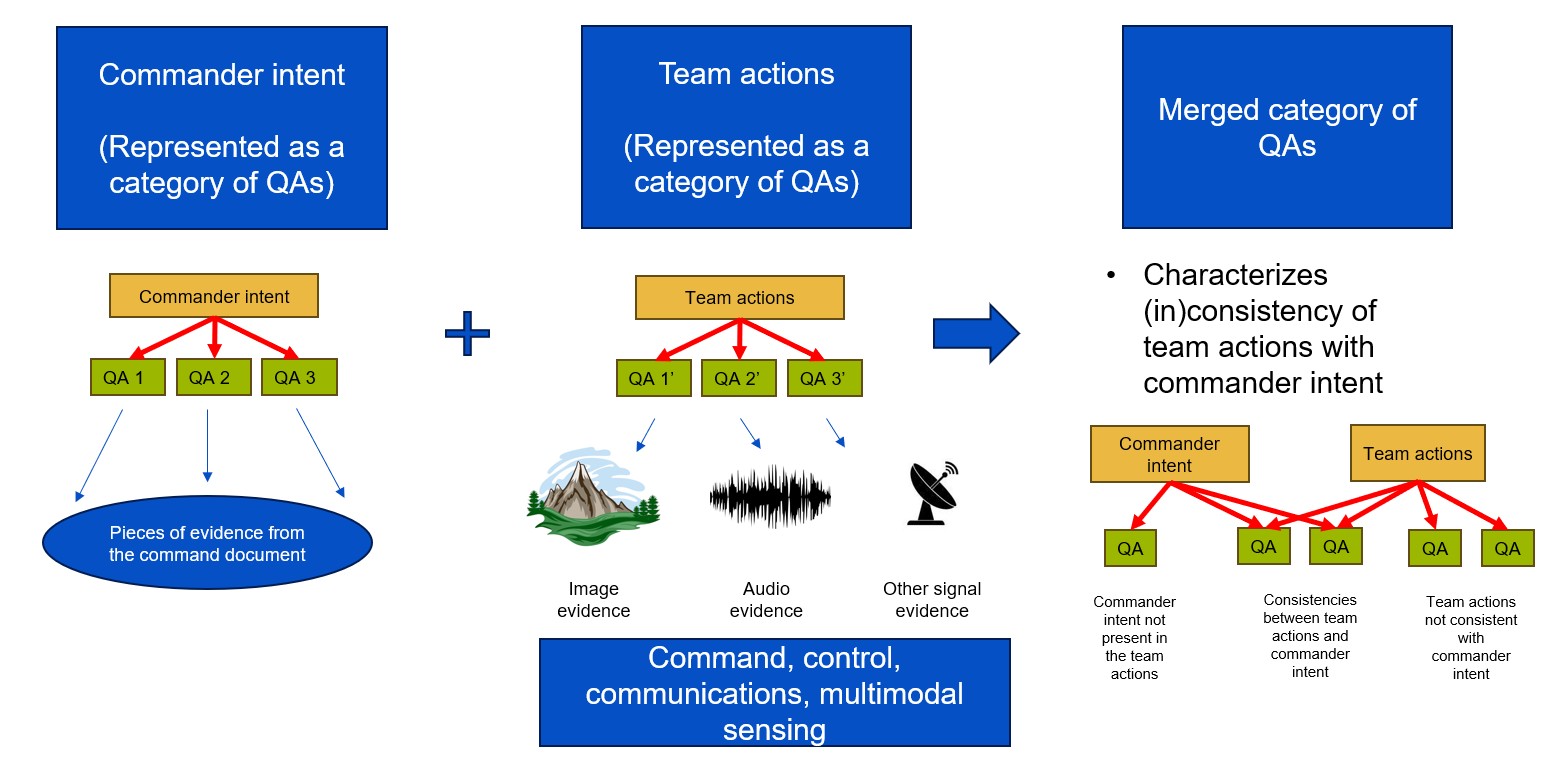}
    \caption{
        The process by which alignment between commander intent and actions on the ground can be evaluated using our multi-modal category theoretic framework.
    }
    \label{fig:cmder-intent-alignment}
\end{figure}

\begin{itemize}
    \item Interactive tools that teach students how to write
        \subitem Structure extraction combined with information theoretic measures gives a document organization framework to the person learning to write. The structure extraction shows the rhetorical structure that the learner has created, while the information theoretic measures shed light on how much detail is extraneous or missing as well as on the overall coherence of the document.
    \item Interactive tools that help fine tuned generation of content such as
    \begin{itemize}
        \item Merging multiple scientific papers into a single coherent document
            \subitem Using the category merger described earlier to merge multiple documents into a single coherent document. The idea would be to first carry out the merger of categories and then use a generative model to generate the merged document.
        \item Generating multiple versions of a document customized to different audiences
            \subitem Another application of generation using categories would be to first extract the structure thus stripping away the style of the document. Then use generative models to generate content with a desired style thus creating the ability to generate multiple versions of the original document with each version having a customized style.
        \item Alignment of Systems with Principles such as Safety, Efficiency, etc.
            \subitem The structure extraction and merging of categories enables assessment of alignment of a certain document or set of documents with another set of documents. Usually alignment is used in the context of conformance to certain ethical principles. The category theoretic framework enables generalization to conformance with any set of principles through fine-grained comparison of the orthogonalized QA pairs. Note that alignment can itself be posed as a question answering task such as ``is the current practice of the system conformant with the principle of harmlessness?''
        \item Alignment of Commander Intent with actions on the ground
            \subitem Figure~\ref{fig:cmder-intent-alignment} shows how we assess such alignment in the context of conformance of certain actions with commander intent. We take each assertion of alignment and assess it against multimodal evidence we get from the actions on the ground. Note that the notion of commander's intent extends beyond military applications to direction of any large organization with a ground operation.
        \item Transfer Learning through exegesis
            \subitem When applied to task descriptions, exegesis can be thought of as transfer learning. Given a description of a task $T_1$, exegesis would lead us to a description of task $T_2$ that has some overlap with $T_1$. When tasks are sufficiently similar, this can be elaboration exegesis, When tasks are sufficiently distinct, this can be superdocument exegesis. Some examples include writing on paper vs. writing on a chalk board, making coffee vs making tea, playing table tennis vs playing tennis and ironing shirts vs ironing pants.
            \subitem In other words, the category theoretic representation of the task $T_1$ gives us a compositional understanding of what is involved in carrying out $T_1$ through the orthogonalization process. In developing a description of task $T_2$, we recognize that some of the components of $T_1$ would be unchanged while some would need to be changed or removed to suit $T_2$. The exegesis process starts from $T_1$ and through the component breakdown, adjusts $T_1$ part by part, and adds parts as necessary to get to task $T_2$. We thus have a systematic way to transfer what was learned from task $T_1$ to task $T_2$. A monolithic representation of $T_1$ would not allow such systematic fine-grained adjustment leading from task $T_1$ to task $T_2$.
    \end{itemize}
\end{itemize}

\subsection{Prompt Science}

Our proposed category theoretic framework provides a common representation for all prompting strategies such as chain of thought, slow and fast thinking, etc. because each of those strategies give rise to prompts that we can treat as documents and then represent as a category as explained earlier. Mapping prompts onto categories gives us the following avenues for further research:
\begin{enumerate}
    \item Use the category to summarize the prompt to reduce computation and, using the rate distortion curve mentioned earlier, develop the right trade-off between accuracy and effort. (This might lead to under-described prompts being filled in by the model’s common sense.  It would make sense to test against multiple large pretrained models to mitigate this.)
    \item Note that the category theoretic representation of prompts enables us to extract the various information theoretic measures described earlier. Furthermore, it also leads to a lattice formulation that in turn leads to principled summarization and exegesis. Our framework thus provides scientific basis for prompt engineering and can thus be termed prompt science.
    \item Use consistency constraints to carry out ``analytical continuations of the prompts'' i.e. dot i's and cross t's in the prompts. Such an approach could be thought of as exegesis resulting in fully consistent extension of the original prompt. The category theoretic representation gives us a systematic way to refine as well as redirect prompts based on user needs.
    \item Note that our approach separates the content or essence of a prompt from its style, allowing us to examine how varying each contributes to the success of the prompt.
    \item Note that we can use the category theoretic approach to merge multiple prompts in a semantic manner. Furthermore, as mentioned in point 3 above, we can also develop self-correction techniques within the prompts using the category theoretic representation's inherent properties such as compositionality.
    \item Note that the exegesis described in 3 above can also be used to extend common sense representations into a more rigorous form, by representing current heuristics in category form and then carrying out lateral and depth wise exegesis. In short, our category theoretic approach provides a way to measure the effort and the performance associated with prompts, as well as a way to systematically extend and modify the prompts to suit user needs.

\end{enumerate}

\section{Related work}

Related work in this area includes empirical approaches to question answering using machine learning (see for example Dalvi~\cite{Dalvi-etal_2021_explaining-answers}), mapping questions onto lattice structures (see Knuth \cite{Knuth_2005_toward-question-asking-machines}, \cite{Knuth_2004_deriving-laws}, \cite{Knuth_2004_intelligent-machines}, \cite{Knuth_2005_lattice-duality}, \cite{Knuth_2004_measuring-questions}, \cite{Knuth_2003_what-is-a-question}, \cite{Knuth_2009_measuring-on-lattices}), and other question-answer formulations using dialogue, inference, logic, etc. Our mathematical formulation goes well beyond the empirical machine learning based approaches reported in the literature in capturing document structure formally. The lattice based modeling of questions by Knuth comes closest to our work. We advance upon that body of work through our large model based decomposition of natural language assertions into QA pairs. Note that such decomposition relies on the implicit ontology learned by the large models and thus far exceeds the scope available to hand-crafted structures such as Knuth's.  That allows us to lean on the LLM's generative power to get at soft predicates, which was not possible before in any automated way. Furthermore, we are the first to incorporate rhetorical structure of a document into a QA structure, which enables us to accommodate both ``single'' questions and overall rhetorical structure. 
Our question answer based proximity or distance metric also enables orthogonalization of QA pairs which is an advancement over prior work. The proposed distance metric also seamlessly leads us to a rate distortion theoretic analysis of summarization which is novel. Our category theoretic representation also enables us to go from summarization to the new problem of content exegesis which offers a rich avenue for further research in transfer learning and systematic knowledge expansion, among others. 
Existing work does offer us the possibility of extending our current work in two ways. First, extend existing work such as Knuth's using our category theoretic formulation. Second, incorporate some of it in our framework. For example, the idea of efficient questions (Shah Nawaz~\cite{Nawaz_2010_topological-structure-of-question-theory}) can be used to understand information gain at a detailed level. Dialog based or dynamics based modeling of question answering~\cite{Minica_2011_thesis}
could be used to add a dynamic aspect to the document categories we have developed so far. We could similarly use the QA-SRL work~\cite{He-etal_2015_question-answer-driven-semantic-role-labeling} to improve the conversion of assertions into QA pairs. The QED work~\cite{Lamm-etal_2021_qed} would suggest that a category theoretic representation of context would be more effective, which is a hypothesis that needs to be verified.
Our category theoretic formulation also enables novel content entropy measures in semantic space, leading to assessment of the information gain provided by a document $D_2$ to the document at hand $D_1$, among other results.


\section{Acknowledgements}

We thank Indranil Sur for sharing a curated list of related work.


\printbibliography

\end{document}